\documentclass[letterpaper]{article}
\usepackage{aaai25}
\usepackage{times}
\usepackage{helvet}
\usepackage{courier}
\usepackage[hyphens]{url}
\usepackage{graphicx}
\usepackage{natbib}
\usepackage{caption}
\usepackage{amsmath,amssymb,amsfonts}
\usepackage{booktabs}
\usepackage{multirow}
\usepackage[table]{xcolor}
\usepackage{adjustbox}
\usepackage{xspace}
\definecolor{improved}{gray}{0.88}
\definecolor{viscue}{RGB}{176,42,42}
\definecolor{thermcol}{RGB}{20,115,105}
\newcommand{\vis}[1]{\textcolor{viscue}{#1}}
\newcommand{\therm}[1]{\textcolor{thermcol}{#1}}
\newcommand{\tgfrow}[3]{\begin{minipage}[c]{0.20\linewidth}\centering
\includegraphics[width=\linewidth]{sfig/#1}\\[1pt]{\scriptsize #2}
\end{minipage}\hfill
\begin{minipage}[c]{0.77\linewidth}{\small #3}\end{minipage}}
\newenvironment{tgfpanel}[1][\textwidth]{\begin{adjustbox}{width=#1,center}
\begin{minipage}{\textwidth}
}{\end{minipage}
\end{adjustbox}
}

\providecommand{\pdfinfo}[1]{}
\newcommand{\method}{TGF\xspace}
\newcommand{\dlc}{Dual-LLM Consensus Judge\xspace}
\newcommand{\lhj}{Lexical-Heuristic Judge\xspace}
\newcommand{\Acc}{\ensuremath{\mathrm{Acc}}\xspace}
\newcommand{\EQ}{\ensuremath{\mathrm{EQ}}\xspace}
\newcommand{\Fc}{\ensuremath{\mathrm{F@C}}\xspace}
\newcommand{\HF}{\ensuremath{\mathrm{HF}}\xspace}
\newcommand{\taubar}{\ensuremath{\bar{\tau}}\xspace}
\newcommand{\phibar}{\ensuremath{\Phi}\xspace}

\newcommand{\dtausub}{\ensuremath{\Delta\tau_{\mathrm{sub}}}\xspace}
\newcommand{\dtaupair}{\ensuremath{\Delta\tau_{\mathrm{pair}}}\xspace}

\title{Right Answer, Wrong Heat: Explanation-Aware Evaluation and Thermal-Grounded Feedback for MLLMs on Infrared Images}

\author{
    Yongsong Huang,
    Xiaofeng Liu,
    Tomo Miyazaki\\
    Yaohou Fan,
    Shinichiro Omachi
}
\affiliations{
    Tohoku University,
    Yale University\\
    (hys, tomo, fan.yaohou.t4, shinichiro.omachi.b5)@tohoku.ac.jp,
    xiaofeng.liu@yale.edu
}

\begin{document}
\maketitle

\begin{abstract}
General-purpose multimodal large language models (MLLMs) are increasingly applied to infrared images, where they are commonly scored by answer accuracy alone. However, a correct answer does not ensure that the model's explanation is grounded in infrared thermal evidence. We introduce an explanation-aware evaluation framework that separates answer correctness, output-level explanation groundedness, and thermal grounding for infrared visual questions. Using a Dual-LLM Consensus Judge with a preliminary human-anchor calibration check, we find that correct answers can still rely on weak or visible-light evidence; withholding the original infrared image and showing only a visible-like rendering erodes thermal grounding with little accuracy change; and this erosion is observed most strongly for more capable models but disappears when infrared remains available. We further propose Thermal-Grounded Feedback (\method), a training-free feedback loop that diagnoses explanation-side failures and revises the explanation while preserving the selected answer. On local paired-input validation, \method improves explanation-side grounding without changing answers. These findings suggest that future trustworthy MLLMs for infrared scene understanding should be evaluated and developed to produce thermally grounded explanations rather than merely accurate answers.
\end{abstract}

\section{Introduction}
\label{sec:introduction}

\begin{figure}[t]
    \centering
    \includegraphics[width=\columnwidth]{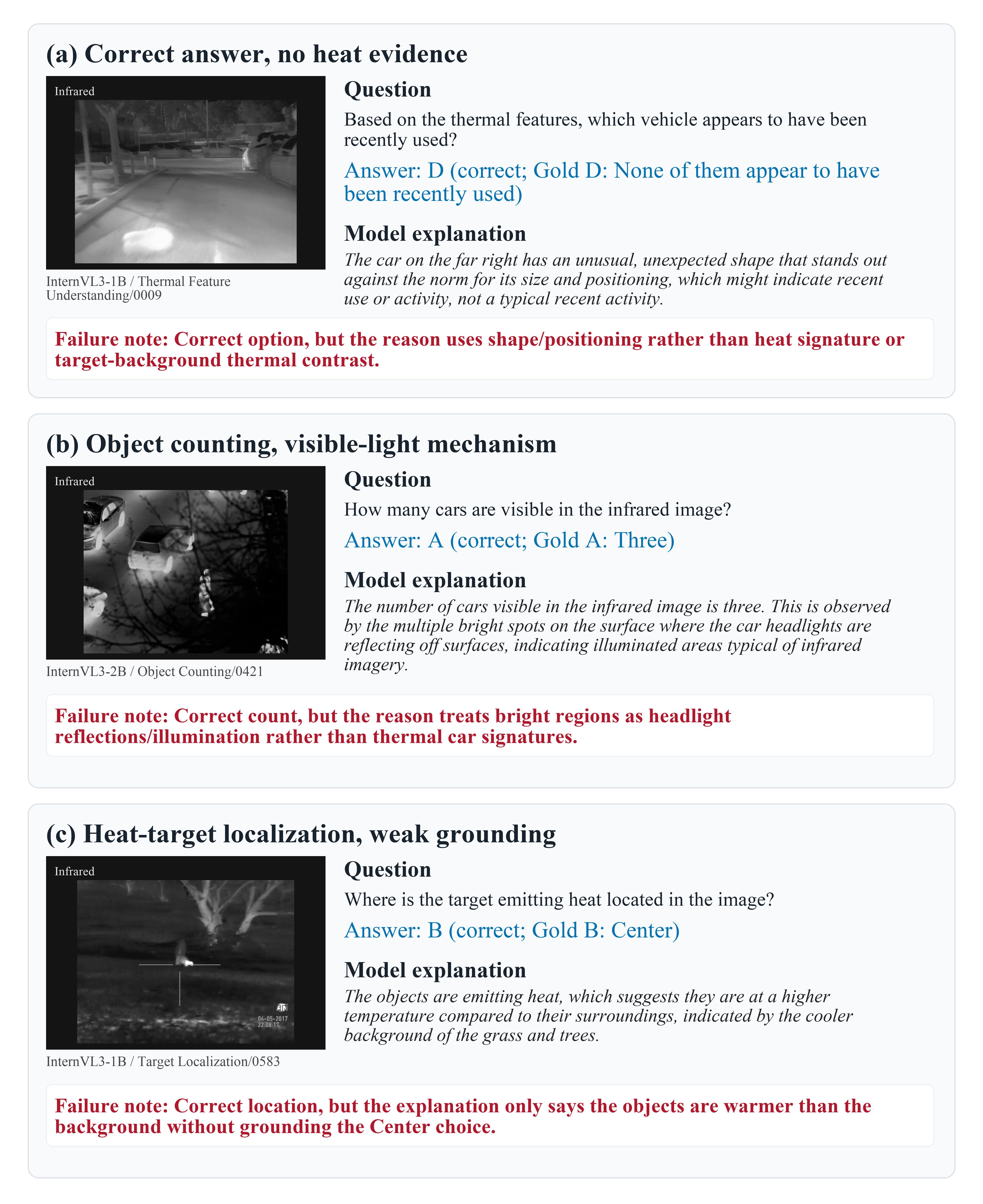}
\caption{\textbf{Right answer, wrong heat.}
Representative infrared questions where the model selects the correct option but justifies it with shape cues, headlight reflections, or weak localization evidence instead of thermal grounding. Answer-only scoring marks these as successes, but the explanation is not grounded in heat. This hidden-failure mode motivates an explanation-aware evaluation that separates being right from being right for thermal reasons.}
    \label{fig:teaser}
\end{figure}

Infrared vision differs from ordinary visible-light perception because its image signal is generated by emitted heat rather than reflected color or texture. This makes thermal imagery valuable for night-time surveillance, autonomous driving, search-and-rescue, industrial inspection, and degraded-visibility scenarios where RGB images fail. It also makes the explanation-grounding problem different: a correct infrared interpretation should rely on relative thermal intensity, heat-signature contours, target--background temperature contrast, grayscale sensor limits, and thermal degradation patterns rather than visible-light color, material, or fine texture. A model that instead defaults to these visible-light cues can land on the correct answer without citing heat evidence at all, a shortcut that answer-only accuracy readily misses.

Language is increasingly used to model, control, and assess infrared imagery. A growing body of work guides infrared--visible fusion with text, semantic priors, or pretrained vision-language models~\cite{wang2025ldfusion,yi2024textif,cheng2025textfusion,li2024texttopixels,zhu2025tesg,qin2026tspfusion,liu2026tsdgfusion,chen2026tofusion,wang2026fdfusion,clipfusion2025}, with additional control from masks, referring expressions, and scene graphs~\cite{sun2026ctrlfuse,zhao2026controlfuse,ma2025risfusion}, while MLLMs are now used to assess perceptual and semantic quality~\cite{you2024depictqa,wang2023clipiqa,cheng2026evanet}. The models actually deployed on infrared tasks, however, are usually general-purpose MLLMs trained predominantly on RGB data. We therefore ask whether such a model, applied to an infrared image, grounds its own explanation in thermal evidence rather than in transferred visible-light priors.

Existing benchmarks cannot settle this question, because they score only the chosen answer: accuracy alone cannot separate a heat-grounded answer from a lucky guess, a dataset prior, or a visible-light shortcut. Figure~\ref{fig:teaser} makes this failure concrete: the model selects the correct option but relies on shape cues, visible-light illumination, or poorly localized thermal evidence. The risk only grows when an infrared image is enhanced or translated into a visible-like representation, which can recover object identity but also invites ordinary RGB semantics and color-like cues that are not valid thermal evidence.

We study this gap by treating the evaluation of general-purpose MLLMs on infrared images as an explanation-aware measurement problem. Instead of scoring only the chosen answer, we ask the model to emit a structured output $o=(a,e,V,u)$ containing an answer $a$, explanation $e$, evidence list $V$, and uncertainty $u$. We then separate three layers: answer correctness $c$, output-level explanation groundedness $\phi$, and thermal grounding $\tau$. This separation yields joint metrics such as explanation-quality-adjusted accuracy ($\EQ=\mathbb{E}[c\phi]$, the share of answers that are both correct and grounded at the output level), grounded-at-correct ($\Fc=\EQ/\Acc$), and hidden-failure rate ($\HF=1-\Fc$). It also lets us test whether thermal grounding changes when the input modality changes.

Our framing is deliberately narrow. We do not measure explanation quality in general, nor calibrate an LLM judge as an end in itself. We ask a domain-specific question: whether the cited evidence constitutes valid \emph{thermal} evidence. We operationalize this question with an output-level grounding rule gated on thermal physics and an input manipulation that withholds the original infrared image and provides only a visible-like rendering, thereby probing the evidence source without changing answer scoring. This separates our problem from generic explanation-groundedness, LLM-judge calibration, and self-revision studies, where valid evidence is not tied to a sensing modality.

This measurement reveals a failure that answer-only scoring cannot detect. A large share of correct IR-only answers are not thermally grounded, and withholding the original infrared image while showing a translated visible rendering erodes thermal grounding while barely changing accuracy. We observe this erosion most strongly for more capable models, which are more likely to express transferred visible-light priors once the original infrared image is unavailable. Yet this capability-associated pattern disappears when infrared remains available alongside the visible view, suggesting that the association is tied to modality substitution rather than RGB augmentation itself. We quantify all three effects in the experiments.

To turn this diagnosis into mitigation, we formulate Thermal-Grounded Feedback (\method) as a training-free, multi-agent feedback loop. A judge agent scores the frozen output, four feedback agents specialize in thermal evidence, evidence grounding, visible-cue quarantine, and uncertainty calibration, a reviser rewrites the explanation, and a selector re-judges candidate revisions before accepting them. \method requires no weight access, fine-tuning, or gold-answer supervision. In local-200 validation, it keeps answer accuracy fixed while improving \Fc (0.875$\to$0.939), \EQ, thermal grounding, and evidence grounding, while cutting parse failure from 0.121 to 0.043. A paired LLM preference evaluation independently favors \method over no-feedback across explanation dimensions.

In summary, our contributions are as follows:
\begin{itemize}
    \item \textbf{Explanation-aware evaluation.} A framework, with a preliminary human-anchor calibration check for the Dual-LLM Consensus Judge, that separates answer correctness from output-level explanation groundedness and thermal grounding, and uses this to expose that general-purpose MLLMs often answer infrared questions correctly while their explanations cite visible-light cues.
    \item \textbf{Thermal-Grounded Feedback.} A training-free, multi-agent procedure that, in the IR+RGB settings where infrared is available yet a third of correct answers stay weakly grounded at the output level, pulls explanations back to thermal evidence without changing answers or accessing weights.
    \item \textbf{Diagnosis and outlook.} We quantify hidden failure, substitution erosion, and an observed capability-associated substitution pattern, and argue that infrared MLLMs should be optimized for thermally grounded output explanations, not answer accuracy alone.
\end{itemize}

\section{Methodology}
\label{sec:methodology}

\subsection{Explanation-Aware Evaluation}
\label{sec:evaluation}

\paragraph{Benchmark and input settings.}
We build our evaluation on IF-Bench~\cite{zhang2025ifbench}, a curated infrared visual-question-answering benchmark with 680 four-option questions over 499 real infrared images. The benchmark covers both explicitly thermal questions and broader scene-understanding questions, which lets us test whether a correct answer is supported by thermal evidence or by visible-light priors that can also solve part of the task.

To probe the source of evidence, we evaluate each model under five input settings that vary how visible-light information is supplied relative to the infrared image. \textbf{IR-only} provides the infrared image alone and is our diagnostic baseline. \textbf{Translated-RGB} \emph{replaces} the infrared image with a visible-like rendering produced by an infrared-to-RGB translation model~\cite{zhang2025ifbench}, withholding the original infrared image. \textbf{IR+prior} adds a short infrared-domain textual hint to the infrared image. \textbf{IR+RGB} pairs the infrared image with its translated rendering, so thermal evidence \emph{remains available}. \textbf{IR+RGB+prior} additionally supplies the textual hint. The contrast between Translated-RGB, which withholds the original infrared image, and the IR+RGB settings, which retain infrared, is central to our analysis: it distinguishes failures caused by adding visible-light information from those caused by \emph{substituting} visible-light information for the original infrared signal.

\paragraph{Task output and notation.}
For each question, a model observes an input setting $s$ and produces a structured output
\begin{equation}
o=(a,e,V,u),
\end{equation}
where $a$ is the selected answer, $e$ is a free-form explanation, $V=\{(t_j,r_j,q_j)\}_{j=1}^m$ is an evidence list of target, region, and thermal cue triples, and $u$ is an uncertainty statement. Let $g$ be the gold answer and let
\begin{equation}
c = \mathbf{1}[a=g]
\end{equation}
denote answer correctness, where $\mathbf{1}[\cdot]$ is the indicator function that equals $1$ when its condition holds and $0$ otherwise.

\paragraph{Output-Level Groundedness and Thermal Grounding.}
We score explanations using three answer-independent components. The thermal-grounding score $\tau\in\{0,1,2,3\}$ measures whether the explanation uses valid infrared evidence. The evidence-grounding score $\epsilon\in\{0,1,2,3\}$ measures whether the stated evidence specifies a target, image region, and thermal cue. The visible-cue misuse flag $h\in\{0,1\}$ marks cases where visible-light evidence such as color or texture is used as support for a thermal conclusion. Let $pf$ denote parse or API failure and $\gamma$ denote whether the question dimension is thermal-gated. We define strict output-level explanation groundedness as
\begin{equation}
\phi = (1-pf)\,\mathbf{1}[\tau\ge2 \vee \gamma=0]\,\mathbf{1}[\epsilon\ge2]\,(1-h).
\end{equation}
The joint grounded-answer indicator is
\begin{equation}
f = c\phi.
\end{equation}
This distinction is important: $\phi$ measures output-level explanation groundedness independent of whether the answer is correct, while $f$ requires both a correct answer and a grounded explanation.

\paragraph{Aggregate metrics.}
For a model-setting bucket, we report answer accuracy $\Acc=\mathbb{E}[c]$, thermal-grounding score $\taubar=\mathbb{E}[\tau]$, output-grounded rate $\phibar=\mathbb{E}[\phi]$, EQ-adjusted accuracy $\EQ=\mathbb{E}[f]=\mathbb{E}[c\phi]$, grounded-at-correct $\Fc=\EQ/\Acc$, and hidden-failure rate $\HF=1-\Fc$.
Because $f_i=c_i\phi_i\le c_i$, $\EQ\le\Acc$ by construction. Therefore, the existence of an answer--EQ gap is not itself a hypothesis test; the meaningful evidence is the magnitude of the gap and its variation across models and settings.

\paragraph{Dual-LLM Consensus Judge.}
The reported evaluator is a \dlc combining GPT-5.5 and Gemini-3.1-Pro judgments. For each output, each judge returns thermal usage, a thermal score, evidence grounding, visible-cue misuse, and a label for output-level explanation groundedness. For groundedness and visible-cue misuse, the consensus retains a label only when both judges agree; for $\tau$, it averages the two thermal scores. We use a 48-item blind human anchor as a preliminary calibration check for the strict rule: across thermal grounding, visible-cue misuse, and strict groundedness, the two strong judges agree with each other at judge--judge $\kappa=0.71$--$0.95$, and the consensus agrees with the human anchor at $\kappa=0.80$--$0.89$ (Table~\ref{tab:judge_validation}). We also score every output with a \lhj (LHJ), a transparent rule-based judge that derives the same labels from regular-expression matching of thermal terms (heat, temperature, contrast) and visible-light terms (color, material, texture), counting a visible cue as misuse only when no visible view is provided. It is cheap and fully reproducible and serves as our baseline, but it agrees far less with the human anchor, because keyword overlap is a weak proxy for output-level thermal grounding, which is why all headline numbers are reported on the \dlc rather than the LHJ.

\begin{table}[t]
\centering
\small
\caption{\textbf{Preliminary human-anchor calibration for the Dual-LLM Consensus Judge.}
Cohen's $\kappa$ across the three labels. The two independent strong judges agree with each other (Judge--Judge, GPT vs.\ Gemini), and the consensus agrees with a 48-item blind human anchor (Human--Consensus). The headline conclusions are therefore reported on the \dlc with this preliminary calibration in mind.}
\label{tab:judge_validation}
\begin{tabular}{lcc}
\toprule
Label & Judge--Judge & Human--Consensus \\
\midrule
Thermal grounding & 0.77 & 0.89 \\
Visible-cue misuse & 0.95 & 0.80 \\
Groundedness (strict) & 0.71 & 0.85 \\
\bottomrule
\end{tabular}
\end{table}

\subsection{Thermal-Grounded Feedback}
\label{sec:tgf}

\begin{figure*}[t]
    \centering
    \includegraphics[width=\linewidth]{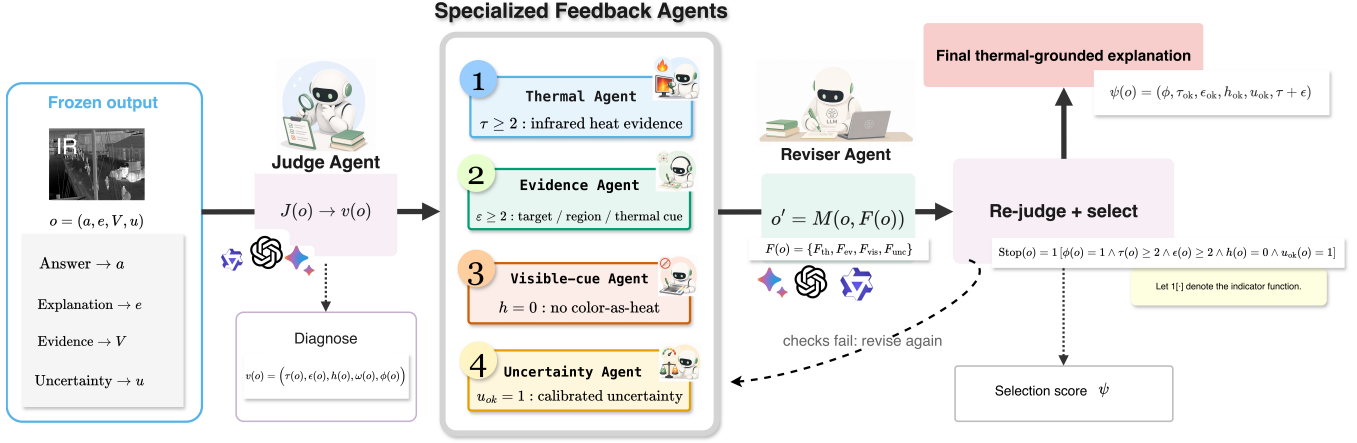}
    \caption{\textbf{Thermal-Grounded Feedback (\method) framework.}
    \method is a multi-agent feedback loop with separated roles over a frozen output $o=(a,e,V,u)$. A judge agent produces a diagnosis $v(o)$ over thermal grounding, evidence grounding, visible-cue misuse, uncertainty calibration, and output-level groundedness. Four feedback agents instantiate $F(o)=\{F_{\mathrm{th}},F_{\mathrm{ev}},F_{\mathrm{vis}},F_{\mathrm{unc}}\}$, a reviser produces $o'=M(o,F(o))$, and a selector re-judges candidate revisions. The selected output maximizes $\psi$, so the loop targets explanation repair without exposing the gold answer, changing the evaluation data, or accessing model weights.}
    \label{fig:tgf_framework}
\end{figure*}

Our evaluation diagnoses a failure but does not fix it: a model can select the correct option yet justify it with visible-light cues instead of thermal evidence. This tendency is most extreme when the original infrared image is withheld and only a visible rendering is shown, but it persists even when infrared remains available alongside a competing visible view, leaving a substantial pool of correct-but-weakly-grounded outputs in exactly the paired settings where a repair can still draw on real thermal evidence. This is an explanation-side failure: the answer may already be correct, whereas the explanation, evidence, or uncertainty may still lack thermal grounding; in principle, these components can be repaired without changing the answer or retraining the model. Retraining is also often impractical: the strongest MLLMs are closed-source, and supervision for infrared explanations is scarce. We therefore introduce \method, a training-free, inference-time procedure that steers a model to re-justify its answer with thermal evidence.

\method is organized around the pipeline in Figure~\ref{fig:tgf_framework}, but the key design choice is not merely to ask the model to ``try again.'' Inspired by agentic restoration systems that separate perception, planning, expert execution, and reflection~\cite{zhu2025agenticir,lin2025jarvisir,jiang2025mair}, we cast explanation repair as a lightweight multi-agent loop over structured text. The agents are functional roles implemented through prompts and structured outputs; they do not require new training data or access to model internals. This framing is useful because the failure we target is compositional: a correct answer can still have missing thermal evidence, vague evidence localization, visible-cue leakage, or miscalibrated uncertainty, and each failure benefits from a different intervention.

The input to \method is a frozen model output $o=(a,e,V,u)$, where the answer $a$ may already be correct but the explanation $e$, evidence list $V$, or uncertainty statement $u$ may be poorly grounded. The first role is the \emph{judge agent}. It maps the output to $v(o)=(\tau(o),\epsilon(o),h(o),u_{\mathrm{ok}}(o),\phi(o))$, where $\tau$ measures thermal grounding, $\epsilon$ measures evidence grounding, $h$ flags visible-cue misuse, $u_{\mathrm{ok}}$ indicates calibrated uncertainty, and $\phi$ is strict output-level explanation groundedness. This step converts an underspecified failure into typed signals that the downstream agents can act on.

The second role is a set of \emph{feedback agents}, one per repair dimension. Folding all repairs into a single instruction to ``be more thermal'' conflates distinct defects and can even trade one for another, for instance, adding heat language while leaving visible-cue leakage in place. Each dimension, therefore, receives its own targeted message derived from the judge's diagnosis. The thermal agent directs the reviser to ground the answer in infrared heat evidence; the evidence agent asks it to name the target, image region, and thermal cue; the visible-cue agent quarantines color, material, or RGB-like descriptions wherever they stand in for thermal evidence; and the uncertainty agent tempers overconfident claims when the thermal evidence is thin. Together, these four role-specific messages define
\begin{equation}
F(o)=\{F_{\mathrm{th}}(o),F_{\mathrm{ev}}(o),F_{\mathrm{vis}}(o),F_{\mathrm{unc}}(o)\}.
\end{equation}

The third role is the \emph{reviser}. It receives the original output and the feedback package and produces
\begin{equation}
o'=M(o,F(o)).
\end{equation}
In the answer-fixed setting used for the main mitigation experiment, the parsed answer is kept equal to the frozen answer so that any improvement is attributable to explanation repair rather than answer switching. Thus, unlike pixel-level restoration agents that invoke image-processing tools, the reviser only edits the explanation, evidence list, and uncertainty statement.

The final role is the \emph{selector}. It re-judges the revised candidate and compares it with previous candidates using the lexicographic selection score
\begin{equation}
\begin{aligned}
\psi(o)=(&\phi(o), \mathbf{1}[\tau(o)\ge2],
          \mathbf{1}[\epsilon(o)\ge2], \\
        &\mathbf{1}[h(o)=0], u_{\mathrm{ok}}(o),
          \tau(o)+\epsilon(o)).
\end{aligned}
\end{equation}
The system keeps the best candidate with the earliest-round tie-breaking. The loop stops when the candidate passes the output-level groundedness rule, is thermally grounded, evidence-grounded, free of visible-cue misuse, and uncertainty-calibrated, or when the maximum number of rounds is reached. In a typical repair, the answer is unchanged while the justification shifts from visible-cue language to relative heat signature and target--background thermal contrast.

This design gives \method three properties needed for a practical mitigation claim: it is training-free and needs no weight access or fine-tuning data; it runs through local inference or external APIs, since the judge and reviser act only on outputs and prompts; and it is evaluated without circularity, because the judge agent inside the loop is never the final evaluation judge.

\section{Experiments}
\label{sec:experiments}

\subsection{Experimental Setup}
We evaluate 11 general-purpose MLLMs, none of them infrared-specialized: Gemini-3.1-Pro, Gemini-3-Flash, GPT-5.5, GPT-5.4, GPT-5.4-mini, Qwen2.5-VL-7B/3B, InternVL3-2B/1B, and InternVL3.5-2B/1B. The benchmark contains 680 infrared multiple-choice items across 10 dimensions. Each model is evaluated under five settings: IR-only, Translated-RGB, IR+prior, IR+RGB, and IR+RGB+prior. The full diagnosis covers all 55 model-setting cells.

The main problem-diagnosis results use the full-data \dlc. The mitigation results use local-200 TGF validation on four local models and two IR-present paired settings. Its before/after table applies the same deterministic rubric as the feedback loop to track repairable changes, while a separate paired LLM preference evaluation provides a semantic check of whether the revised explanations are actually preferred over no-feedback outputs.

\begin{table*}[t]
\centering
\scriptsize
\setlength{\tabcolsep}{1.55pt}
\caption{\textbf{Model-level results across all input settings.}
Full-data \dlc results for IR-only and the four non-baseline settings. Shaded cells indicate improvement over the same model's IR-only value. The complete setting table shows both sides of the argument: replacing IR with translated RGB sharply reduces thermal grounding for capable models, whereas IR-present paired settings often improve \Fc and \EQ without reproducing the substitution collapse. $\tau$ is on a $0$--$3$ scale; all other metrics are in $[0,1]$.}
\label{tab:setting_shift}
\begin{adjustbox}{width=0.85\textwidth}
\begin{tabular}{lcccccccccccccccccccc}
\toprule
\multirow{2}{*}{Model} & \multicolumn{4}{c}{IR-only} & \multicolumn{4}{c}{Translated-RGB} & \multicolumn{4}{c}{IR+prior} & \multicolumn{4}{c}{IR+RGB} & \multicolumn{4}{c}{IR+RGB+prior} \\
\cmidrule(lr){2-5}\cmidrule(lr){6-9}\cmidrule(lr){10-13}\cmidrule(lr){14-17}\cmidrule(lr){18-21}
& Acc & $\tau$ & F@C & EQ & Acc & $\tau$ & F@C & EQ & Acc & $\tau$ & F@C & EQ & Acc & $\tau$ & F@C & EQ & Acc & $\tau$ & F@C & EQ \\
\midrule
Gemini-3.1-Pro & 0.862 & 1.666 & 0.824 & 0.710 & 0.807 & 0.588 & 0.754 & 0.609 & \cellcolor{improved}0.869 & \cellcolor{improved}1.681 & \cellcolor{improved}0.838 & \cellcolor{improved}0.728 & \cellcolor{improved}0.863 & \cellcolor{improved}1.685 & \cellcolor{improved}0.879 & \cellcolor{improved}0.759 & 0.862 & \cellcolor{improved}1.667 & \cellcolor{improved}0.894 & \cellcolor{improved}0.771 \\
Gemini-3-Flash & 0.862 & 2.231 & 0.894 & 0.771 & 0.812 & 0.987 & 0.818 & 0.665 & \cellcolor{improved}0.866 & \cellcolor{improved}2.308 & \cellcolor{improved}0.902 & \cellcolor{improved}0.781 & \cellcolor{improved}0.868 & 2.169 & \cellcolor{improved}0.971 & \cellcolor{improved}0.843 & 0.859 & 2.176 & \cellcolor{improved}0.971 & \cellcolor{improved}0.834 \\
GPT-5.5 & 0.828 & 2.160 & 0.957 & 0.793 & 0.770 & 1.077 & 0.920 & 0.708 & 0.824 & \cellcolor{improved}2.173 & \cellcolor{improved}0.964 & \cellcolor{improved}0.794 & \cellcolor{improved}0.837 & \cellcolor{improved}2.163 & \cellcolor{improved}0.988 & \cellcolor{improved}0.826 & \cellcolor{improved}0.841 & 2.149 & \cellcolor{improved}0.988 & \cellcolor{improved}0.831 \\
GPT-5.4 & 0.790 & 2.207 & 0.948 & 0.749 & 0.749 & 0.919 & 0.850 & 0.637 & \cellcolor{improved}0.806 & 2.182 & \cellcolor{improved}0.958 & \cellcolor{improved}0.772 & \cellcolor{improved}0.821 & 1.973 & \cellcolor{improved}0.950 & \cellcolor{improved}0.779 & \cellcolor{improved}0.815 & 2.043 & \cellcolor{improved}0.966 & \cellcolor{improved}0.787 \\
GPT-5.4-mini & 0.747 & 1.929 & 0.890 & 0.665 & 0.726 & 1.020 & 0.880 & 0.639 & 0.746 & \cellcolor{improved}1.964 & \cellcolor{improved}0.925 & \cellcolor{improved}0.690 & 0.745 & 1.810 & \cellcolor{improved}0.947 & \cellcolor{improved}0.705 & \cellcolor{improved}0.754 & 1.884 & \cellcolor{improved}0.965 & \cellcolor{improved}0.727 \\
Qwen2.5-VL-7B & 0.710 & 1.340 & 0.679 & 0.482 & \cellcolor{improved}0.717 & 1.001 & \cellcolor{improved}0.769 & \cellcolor{improved}0.551 & 0.703 & \cellcolor{improved}1.401 & 0.674 & 0.474 & \cellcolor{improved}0.741 & 1.214 & 0.668 & \cellcolor{improved}0.495 & \cellcolor{improved}0.737 & 1.312 & \cellcolor{improved}0.715 & \cellcolor{improved}0.526 \\
Qwen2.5-VL-3B & 0.638 & 1.167 & 0.529 & 0.337 & \cellcolor{improved}0.665 & 0.841 & \cellcolor{improved}0.577 & \cellcolor{improved}0.384 & \cellcolor{improved}0.643 & \cellcolor{improved}1.286 & \cellcolor{improved}0.577 & \cellcolor{improved}0.371 & \cellcolor{improved}0.682 & 1.037 & \cellcolor{improved}0.534 & \cellcolor{improved}0.365 & \cellcolor{improved}0.682 & \cellcolor{improved}1.209 & \cellcolor{improved}0.575 & \cellcolor{improved}0.393 \\
InternVL3-2B & 0.669 & 0.950 & 0.376 & 0.251 & 0.641 & 0.547 & \cellcolor{improved}0.457 & \cellcolor{improved}0.293 & 0.659 & \cellcolor{improved}1.149 & \cellcolor{improved}0.500 & \cellcolor{improved}0.329 & 0.650 & 0.808 & \cellcolor{improved}0.425 & \cellcolor{improved}0.276 & 0.663 & 0.944 & \cellcolor{improved}0.449 & \cellcolor{improved}0.297 \\
InternVL3.5-2B & 0.649 & 1.050 & 0.512 & 0.332 & \cellcolor{improved}0.688 & 0.700 & \cellcolor{improved}0.578 & \cellcolor{improved}0.398 & 0.643 & \cellcolor{improved}1.143 & \cellcolor{improved}0.554 & \cellcolor{improved}0.356 & \cellcolor{improved}0.669 & 0.976 & \cellcolor{improved}0.554 & \cellcolor{improved}0.371 & \cellcolor{improved}0.660 & \cellcolor{improved}1.113 & \cellcolor{improved}0.586 & \cellcolor{improved}0.387 \\
InternVL3.5-1B & 0.576 & 0.553 & 0.207 & 0.119 & \cellcolor{improved}0.581 & 0.283 & \cellcolor{improved}0.440 & \cellcolor{improved}0.256 & 0.569 & \cellcolor{improved}0.688 & \cellcolor{improved}0.271 & \cellcolor{improved}0.154 & \cellcolor{improved}0.609 & 0.464 & \cellcolor{improved}0.246 & \cellcolor{improved}0.150 & \cellcolor{improved}0.621 & \cellcolor{improved}0.574 & \cellcolor{improved}0.280 & \cellcolor{improved}0.174 \\
InternVL3-1B & 0.543 & 0.603 & 0.122 & 0.066 & \cellcolor{improved}0.546 & 0.314 & \cellcolor{improved}0.226 & \cellcolor{improved}0.124 & 0.525 & \cellcolor{improved}0.693 & \cellcolor{improved}0.132 & \cellcolor{improved}0.069 & \cellcolor{improved}0.561 & 0.484 & \cellcolor{improved}0.134 & \cellcolor{improved}0.075 & \cellcolor{improved}0.553 & 0.505 & \cellcolor{improved}0.144 & \cellcolor{improved}0.080 \\
\midrule
Mean & 0.716 & 1.441 & 0.631 & 0.480 & 0.700 & 0.752 & \cellcolor{improved}0.661 & 0.479 & 0.714 & \cellcolor{improved}1.515 & \cellcolor{improved}0.663 & \cellcolor{improved}0.502 & \cellcolor{improved}0.731 & 1.344 & \cellcolor{improved}0.663 & \cellcolor{improved}0.513 & \cellcolor{improved}0.732 & 1.416 & \cellcolor{improved}0.685 & \cellcolor{improved}0.528 \\
\bottomrule
\end{tabular}
\end{adjustbox}
\end{table*}

\subsection{Answer Accuracy Hides Weak Output-Level Grounding}

\begin{figure}[tb]
    \centering
    \includegraphics[width=0.85\columnwidth]{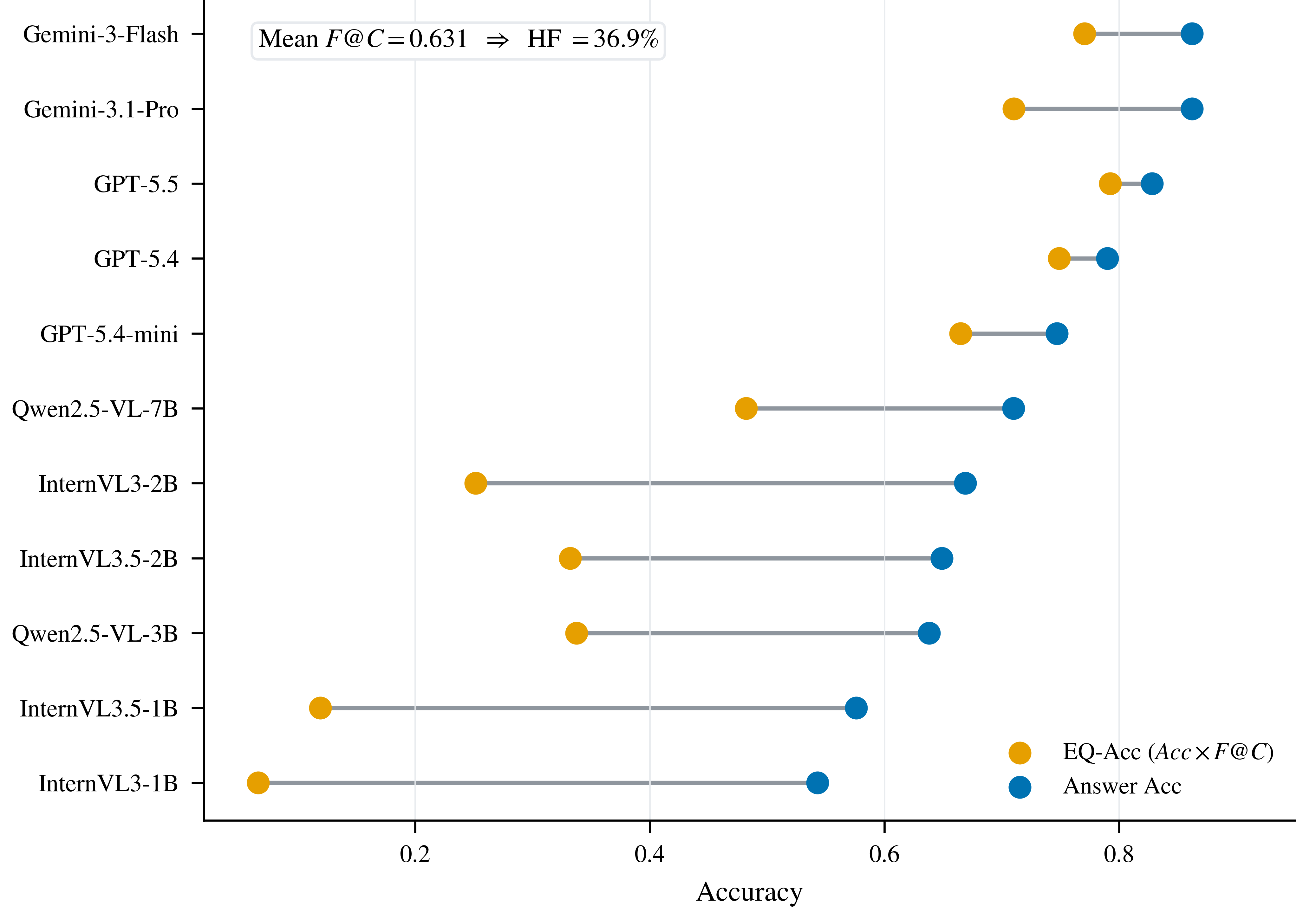}
    \caption{\textbf{Answer accuracy overstates output-level infrared grounding.}
    For each model, EQ-adjusted accuracy is computed as $\Acc\times\Fc$, so it counts only answers that are both correct and grounded at the output level. The mean full-data consensus $\Fc$ is 0.631, giving a hidden-failure rate of 36.9\%.}
    \label{fig:hidden_failure}
\end{figure}

Answer accuracy substantially overstates infrared understanding relative to what the explanations support (Figure~\ref{fig:hidden_failure}). Under the final \dlc, the mean IR-only $\Fc$ is only 0.631, so $\HF=36.9\%$ of correct answers fail the output-level grounding criterion. This rate far exceeds the \lhj estimate, because matching thermal keywords is a weak proxy for output-level thermal grounding. The gap is also highly uneven across models: $\Fc$ ranges from about 0.82 to 0.96 for the strongest models but falls to 0.12--0.53 for the weakest, so hidden failure is capability-bimodal rather than a uniform offset.

These hidden failures are primarily about cited thermal evidence, not formatting noise. Among the IR-only correct-but-weakly-grounded cases, the consensus judge marks $72\%$ as not grounded in thermal evidence and $46\%$ as relying on a visible-light cue (color, material, texture) as evidence, whereas parse or API failures account for only $0.1\%$; their mean thermal-grounding score is $1.74$, below the $\tau\ge2$ bar required for the strict grounding rule. Hidden failure is therefore a wrong-heat problem rather than a wrong-format one: the answer is right, but the stated reason is visible-light or thermally shallow.

\begin{figure}[tb]
    \centering
    \includegraphics[width=0.85\columnwidth]{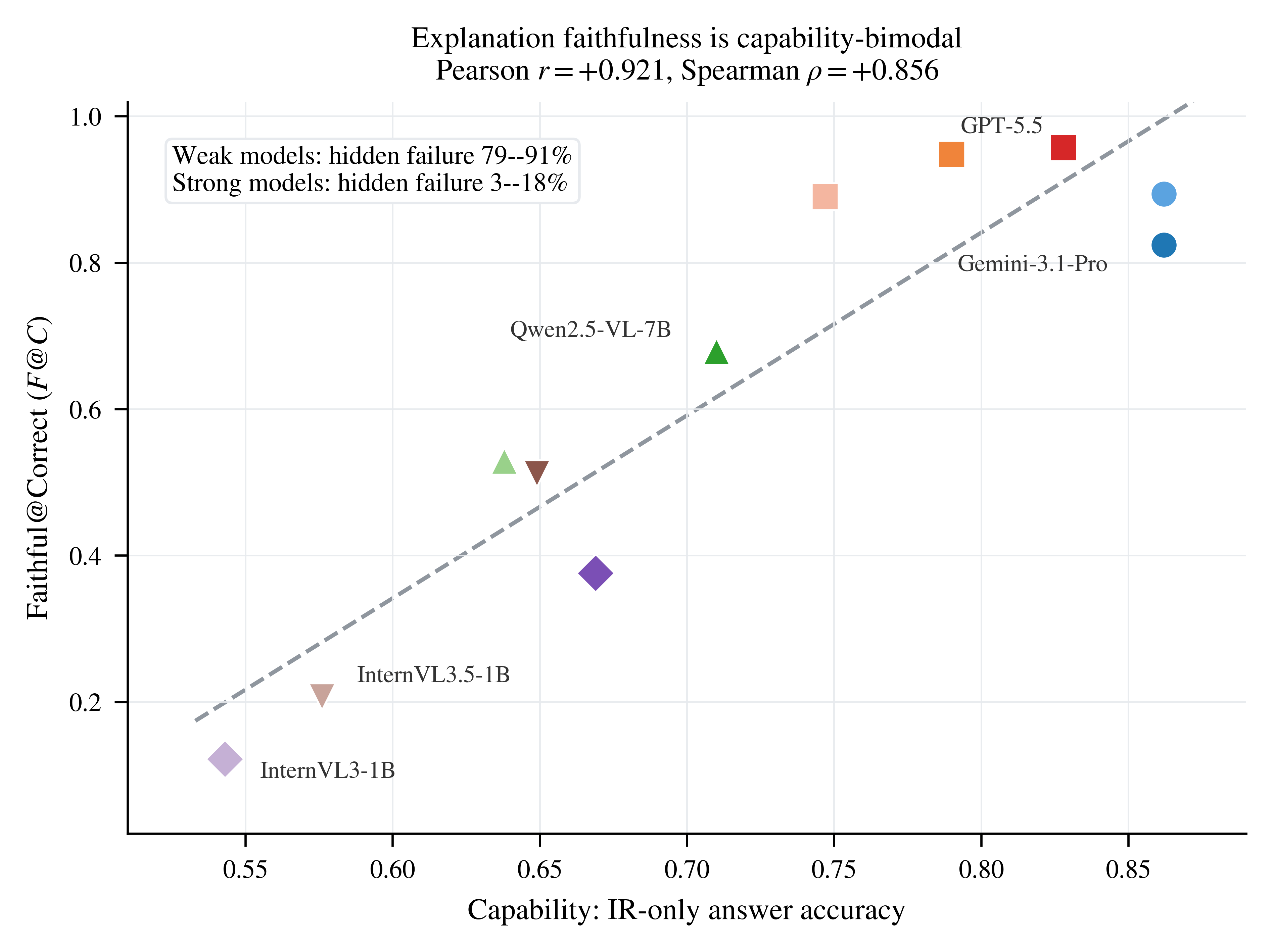}
    \caption{\textbf{Output-level groundedness itself scales with model capability.}
    Full-data consensus $\Fc$ increases with IR-only answer accuracy (Pearson $r=+0.921$, Spearman $\rho=+0.856$), indicating that answer ability and output-level explanation groundedness are related but not identical axes.}
    \label{fig:faithful_capability}
\end{figure}

Output-level groundedness, like accuracy, climbs with capability (Figure~\ref{fig:faithful_capability}), yet the two axes are far from interchangeable: even the strongest models retain a residual pool of hidden failures that accuracy alone cannot localize. A capability-driven groundedness trend therefore reinforces, rather than removes, the case for explanation-aware evaluation.

Hidden failure also concentrates on the dimensions a visible-light prior can already answer. The highest rates fall on semantic dimensions that do not strictly require heat, such as scene understanding (HF $49\%$ at $0.92$ accuracy), commonsense reasoning ($47\%$), and viewpoint of capture ($43\%$), where the correct option is recoverable from layout or context rather than thermal evidence. The lowest rates belong to the genuinely thermal dimensions, thermal feature understanding ($13\%$) and thermal feature reasoning ($16\%$), which also carry the highest consensus thermal score ($\bar\tau\approx2.3$). Even though these two dimensions are held to the stricter requirement $\tau\ge2$, they still fail least often, while the ungated semantic dimensions fail most. Table~\ref{tab:dim_hidden} makes this dissociation explicit: the 36.1-point spread between scene understanding and thermal feature understanding shows that hidden failure is task-dependent rather than a uniform model offset.

\begin{table*}[t]
\centering
\small
\caption{\textbf{Per-dimension hidden failure on IR-only.}
Full-data \dlc results pooled over all 11 models and sorted by hidden-failure rate \HF{} ($\HF=1-\Fc$). Here $\Phi$ is the strict output-grounded rate and $\bar\tau\in[0,3]$ the thermal-grounding score.}
\label{tab:dim_hidden}
\begin{tabular}{lrccccc}
\toprule
Dimension & $n$ & \Acc & $\bar\tau$ & $\Phi$ & \Fc & \HF \\
\midrule
Scene Understanding & 858 & 0.924 & 1.186 & 0.481 & 0.508 & 0.492 \\
Commonsense Reasoning & 681 & 0.793 & 1.024 & 0.448 & 0.531 & 0.469 \\
Viewpoint of Capture & 638 & 0.677 & 0.737 & 0.390 & 0.569 & 0.431 \\
Action Recognition & 715 & 0.740 & 1.192 & 0.497 & 0.609 & 0.391 \\
Image Theme & 814 & 0.892 & 1.356 & 0.602 & 0.636 & 0.364 \\
Spatial Relationship Understanding & 594 & 0.519 & 1.256 & 0.449 & 0.740 & 0.260 \\
Target Localization & 803 & 0.685 & 1.443 & 0.603 & 0.749 & 0.251 \\
Object Counting & 957 & 0.506 & 1.534 & 0.526 & 0.783 & 0.217 \\
Thermal Feature Reasoning & 649 & 0.797 & 2.304 & 0.710 & 0.843 & 0.157 \\
Thermal Feature Understanding & 770 & 0.616 & 2.301 & 0.579 & 0.869 & 0.131 \\
\bottomrule
\end{tabular}
\end{table*}

\subsection{Translated Visible Input Erodes Thermal Grounding}

When the original infrared image is withheld and only a translated visible rendering is shown, mean thermal grounding drops sharply from 1.441 to 0.752 ($\dtausub=-0.689$) across all 11 models. Answer accuracy changes only from 0.716 to 0.700, and uncertainty does not rise meaningfully. Table~\ref{tab:setting_shift} shows that the same collapse does not characterize the IR-present settings: IR+prior has the highest mean thermal score (1.515), while paired IR+RGB+prior raises the mean \Fc from 0.631 to 0.685 and \EQ from 0.480 to 0.528. The substitution pattern is straightforward: the model keeps answering, but the explanation shifts away from thermal evidence. This is why a generic groundedness or accuracy check is insufficient; the evaluator should know whether the cited evidence is thermal. The erosion is itself dimension-dependent. It is largest for spatial and counting dimensions, such as object counting ($\dtausub=-0.89$), target localization ($-0.83$), and spatial relationship ($-0.78$), where a visible rendering most readily invites layout-based shortcuts, and smallest for the intrinsically thermal dimensions (thermal feature reasoning $-0.32$). That even the thermal dimensions erode is consistent with substitution weakening output-level grounding across the board rather than only on easily substituted tasks.

\subsection{Capability Paradox as an Observed Modality-Substitution Association}

\begin{figure}[t]
    \centering
    \includegraphics[width=\columnwidth]{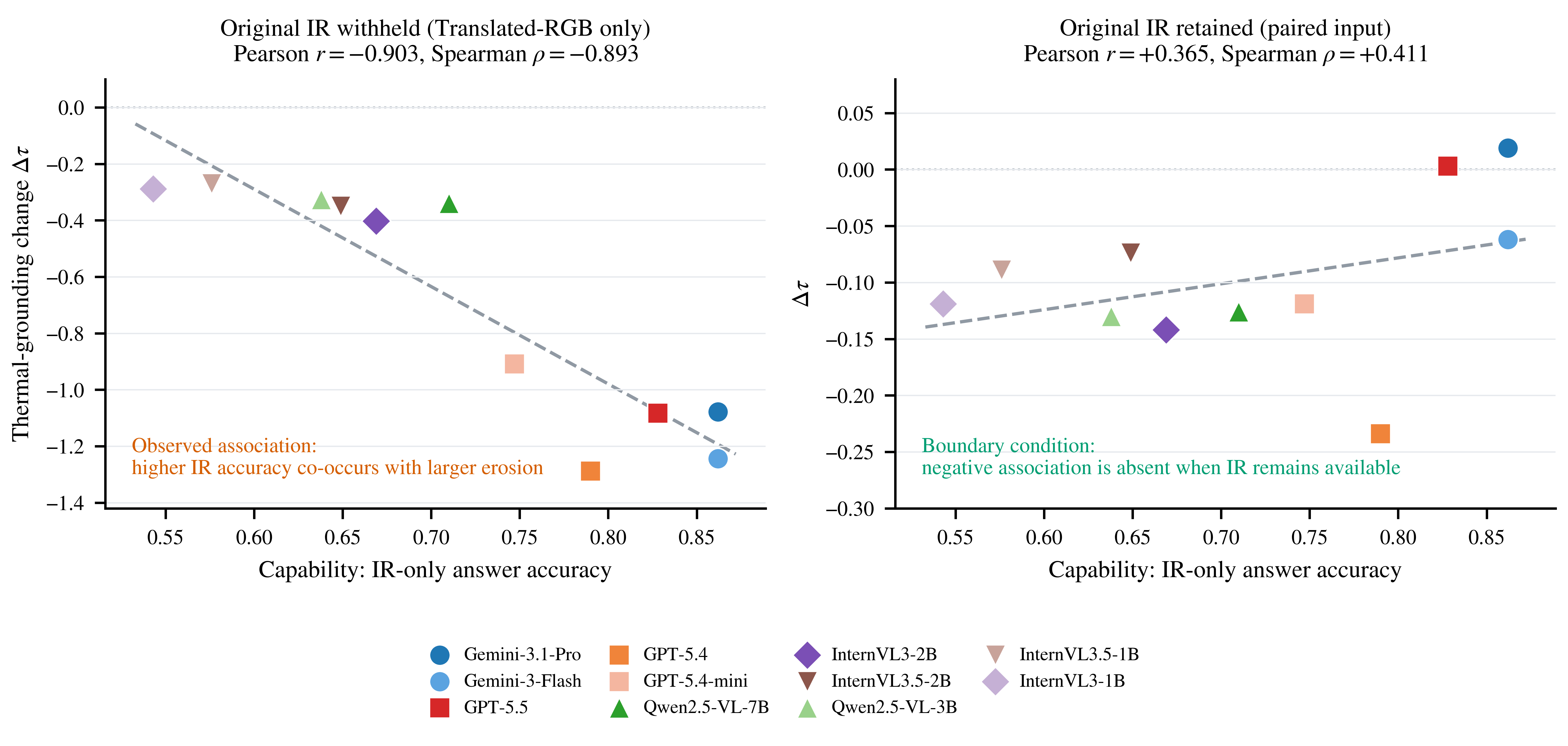}
    \caption{\textbf{The capability-associated gap is bounded by infrared availability.}
    Left: across the evaluated models, higher IR-only accuracy is associated with a larger thermal-grounding decrease when the original infrared image is replaced by a generated visible-like rendering ($r=-0.903$, $\rho=-0.893$). Right: the negative association is absent when infrared remains available in paired IR+RGB input ($r=+0.365$, $\rho=+0.411$). This is a descriptive cross-model association and does not establish an independent causal effect of model capability.}
    \label{fig:capability_paradox}
\end{figure}

\begin{table}[t]
\centering
\small
\caption{\textbf{Per-model thermal-grounding changes from the unified full-data table.}
$\dtausub$ is Translated-RGB minus IR-only, where the generated visible-like rendering replaces infrared. $\dtaupair$ is IR+RGB minus IR-only, where infrared remains available. Models are sorted by IR-only accuracy.}
\label{tab:capability_delta}
\begin{tabular}{lccc}
\toprule
Model & IR \Acc & \dtausub & \dtaupair \\
\midrule
Gemini-3.1-Pro & 0.862 & $-1.078$ & $+0.019$ \\
Gemini-3-Flash & 0.862 & $-1.244$ & $-0.062$ \\
GPT-5.5 & 0.828 & $-1.083$ & $+0.003$ \\
GPT-5.4 & 0.790 & $-1.288$ & $-0.234$ \\
GPT-5.4-mini & 0.747 & $-0.909$ & $-0.119$ \\
Qwen2.5-VL-7B & 0.710 & $-0.339$ & $-0.126$ \\
InternVL3-2B & 0.669 & $-0.403$ & $-0.142$ \\
InternVL3.5-2B & 0.649 & $-0.350$ & $-0.074$ \\
Qwen2.5-VL-3B & 0.638 & $-0.326$ & $-0.130$ \\
InternVL3.5-1B & 0.576 & $-0.270$ & $-0.089$ \\
InternVL3-1B & 0.543 & $-0.289$ & $-0.119$ \\
\bottomrule
\end{tabular}
\end{table}

Across the evaluated models, higher IR-only accuracy is associated with a larger thermal-grounding decrease when the original infrared image is replaced by a generated visible-like rendering ($r=-0.903$, $\rho=-0.893$; Figure~\ref{fig:capability_paradox}). Table~\ref{tab:capability_delta} exposes the model-level pattern behind this aggregate: all 11 models lose thermal grounding under substitution, with the five strongest models dropping by $0.909$--$1.288$ points, compared with $0.270$--$0.403$ for the six open-source models. When infrared remains available, the paired-input changes are much closer to zero ($+0.019$ to $-0.234$) and do not reproduce the same capability ordering. The boundary is therefore visible model by model rather than being an artifact of averaging.

The association remains descriptive and partially baseline-coupled. On the same full-data table, IR-only accuracy correlates strongly with the IR-only thermal score ($r=+0.912$), while the IR-only thermal score correlates with the substitution delta ($r=-0.928$), which mechanically contains the IR-only baseline. Restricting the analysis to the six open-source models weakens the association to $r=-0.742$ ($\rho=-0.771$), and controlling for a closed-source indicator gives a partial correlation of $-0.448$. Within-family comparisons retain the direction but are small, particularly for Qwen2.5-VL. We therefore interpret the result as a capability-associated stress-test pattern with an explicit IR-present boundary, not as evidence that capability independently causes visible-evidence substitution.

\subsection{TGF Mitigates Explanation-Side Failures}

Having characterized the failure, we evaluate \method on the local-200 subset with four local models in the two paired settings (IR+RGB and IR+RGB+prior). These are the target settings: infrared evidence remains available beside a competing visible-like view, while correct answers still have full-data hidden-failure rates of 33.7\% and 31.5\%, respectively. Holding the answer fixed isolates explanation repair from answer switching.

\begin{table}[t]
\centering
\small
\caption{\textbf{\method improves explanation-side metrics on both paired settings.}
Local-200 means over four local models; entries are no-feedback$\to$\method. Answers are fixed.}
\label{tab:tgf_before_after}
\begin{tabular}{lcc}
\toprule
Metric & IR+RGB & IR+RGB+prior \\
\midrule
F@C $\uparrow$ & $0.878\to0.950$ & $0.872\to0.929$ \\
EQ $\uparrow$ & $0.510\to0.546$ & $0.501\to0.534$ \\
Thermal score $\tau$ $\uparrow$ & $2.298\to2.359$ & $2.558\to2.594$ \\
Evidence score $\uparrow$ & $2.843\to2.945$ & $2.845\to2.929$ \\
Parse failure $\downarrow$ & $0.121\to0.039$ & $0.120\to0.048$ \\
\bottomrule
\end{tabular}
\end{table}

Table~\ref{tab:tgf_before_after} shows consistent mean improvements in F@C, EQ, thermal grounding, and evidence grounding, while parse failures fall by more than half. Gains concentrate in repairable weaker-model cells, especially InternVL3-2B, while Qwen cells are already near the ceiling. A paired LLM preference evaluation across 1,600 comparisons independently favors \method over no-feedback for thermal grounding (0.320 vs.\ 0.133), output-level groundedness (0.245 vs.\ 0.138), evidence grounding (0.239 vs.\ 0.128), and overall quality (0.381 vs.\ 0.197). Uncertainty shows no significant preference (0.101 vs.\ 0.106). Across the same explanations, \method makes $5.1\%$ pass the strict grounding criterion that did not before and harms $1.6\%$, with $93.3\%$ unchanged.

\begin{figure}[t]
\centering
\begin{tgfpanel}[\columnwidth]
\tgfrow{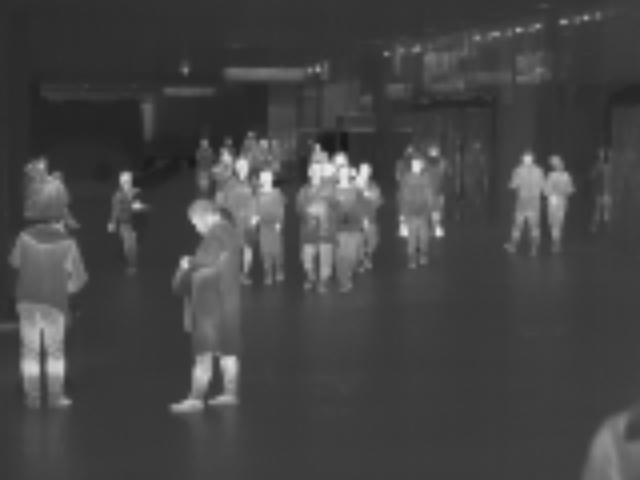}{Object Counting}{\textbf{Q:} How many people are visible? \textbf{(Gold A: 20--30.)}\\[1pt]
\textbf{Before:} ``The image is marked as \vis{`ir\_plus\_translated\_rgb'}, suggesting it uses infrared technology\ldots the count ranges between 20 and 30.''\\[1pt]
\textbf{After:} ``\ldots the group of people near the center appears \therm{consistent in heat signature}\ldots the number ranges between 20 and 30, supported by the \therm{thermal uniformity} observed.''}

\vspace{6pt}
\tgfrow{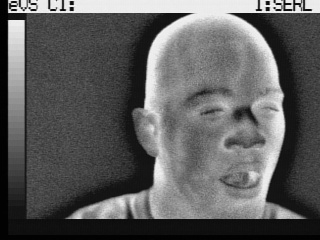}{Thermal Feature Understanding}{\textbf{Q:} Which part of the face shows the lowest temperature? \textbf{(Gold C: nose.)}\\[1pt]
\textbf{Before:} ``\ldots areas with lower temperature appear darker because \vis{infrared light penetrates the skin} more when temperature is low\ldots''\\[1pt]
\textbf{After:} ``\ldots areas with lower temperature appear darker due to the \therm{heat signature}; the nose shows a darker shade than the forehead, cheek, and ears, indicating the \therm{lowest temperature}.''}

\vspace{6pt}
\tgfrow{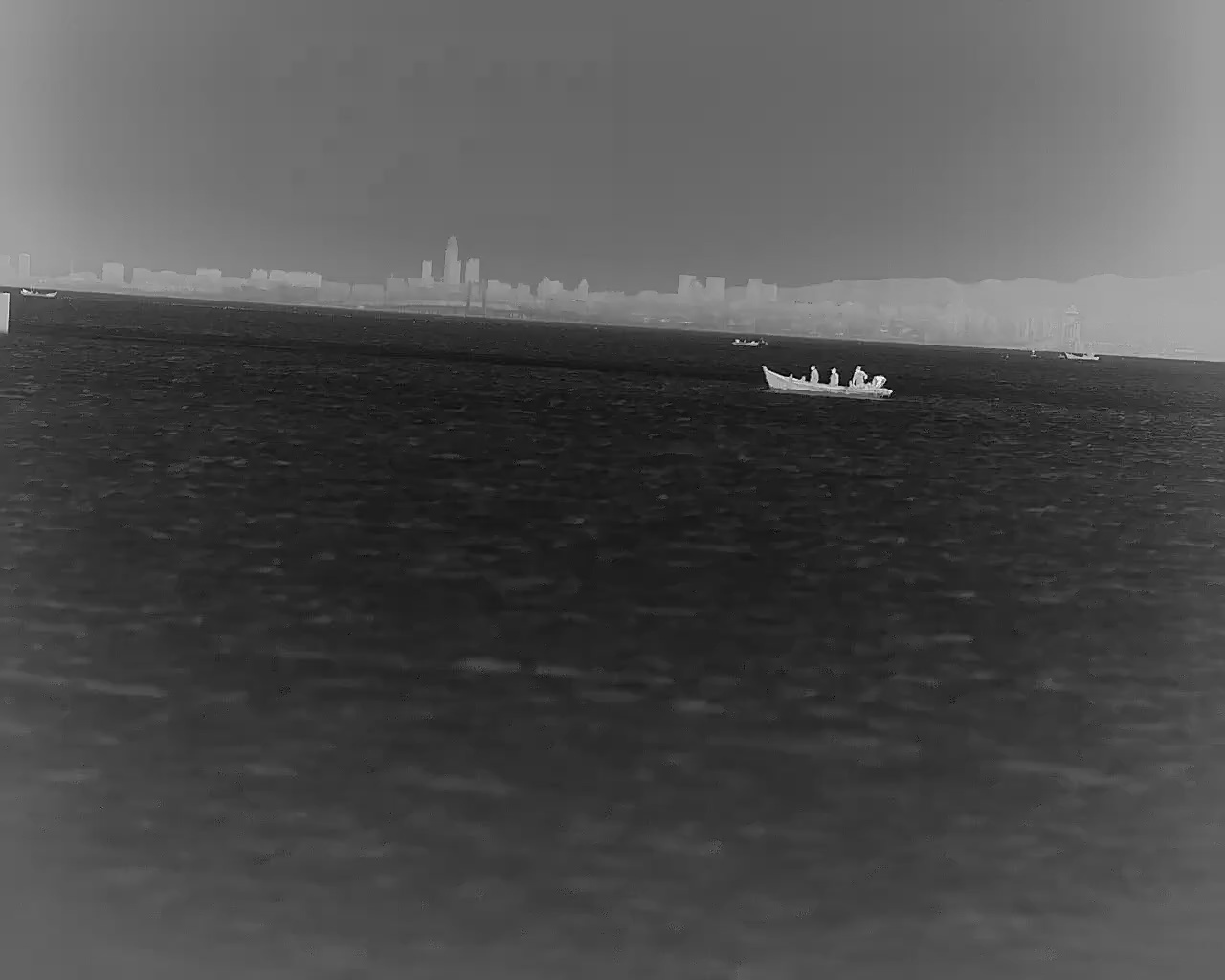}{Thermal Feature Understanding}{\textbf{Q:} Which area has higher thermal activity? \textbf{(Gold D: the boats.)}\\[1pt]
\textbf{Before:} ``The boats emit more heat\ldots \vis{the boats are cooler than the water due to heat absorption}\ldots'' (internally contradictory)\\[1pt]
\textbf{After:} ``The boats have a \therm{higher thermal signature} than the water and distant cityscape, indicated by their \therm{brightness and clarity against the cooler water}.''}
\end{tgfpanel}
\caption{\textbf{\method repairs explanation-side failures while keeping the answer fixed.} The three examples show reliance on an input-format tag, incorrect infrared physics, and an internally contradictory thermal claim; \method replaces each faulty justification with explicit thermal evidence. Examples are quoted verbatim from the Local-200 logs (InternVL3-2B).}
\label{fig:tgf_body}
\end{figure}

Figure~\ref{fig:tgf_body} illustrates the corresponding repairs. \method replaces an input-format tag, incorrect infrared physics, and an internal thermal contradiction with relative heat signatures, thermal uniformity, and target--background contrast, without changing the answer.

\section{Conclusion}
\label{sec:conclusion}

Evaluating general-purpose MLLMs on infrared images should distinguish selecting the right option from justifying it with the right thermal evidence. We propose an explanation-aware framework that separates answer correctness, output-level explanation groundedness, and thermal grounding, scored by a Dual-LLM Consensus Judge checked against a preliminary human anchor. It exposes hidden failures that answer-only scoring conceals, shows that thermal grounding erodes under visible-like modality substitution, and reveals an observed capability-associated pattern that vanishes once infrared remains available. Building on this diagnosis, \method, a training-free feedback loop, repairs a subset of these explanation-side failures without retraining or weight access. We thus recommend reporting these three quantities separately and treating inference-time feedback as a practical lever. These results point future development toward MLLMs that better align their output explanations with infrared evidence, rather than transferred visible-light habits.

\bibliography{aaai25}

\end{document}